\documentclass{article}
\usepackage{spconf,amsmath,graphicx}

\usepackage{algorithmicx}
\usepackage{algpseudocode} % This is REQUIRED to use \begin{algorithmic}

\usepackage[caption=false,font=footnotesize]{subfig}

\renewcommand{\vec}[1]{\boldsymbol{#1}}

\title{Instance Flow Based Online Multiple Object Tracking}
\name{Sebastian Bullinger, Christoph Bodensteiner, Michael Arens}
\address{Fraunhofer Institute of Optronics, System Technologies and Image Exploitation}
\usepackage{tikz}

\newcommand \copyrighttextICIPTwentySeventeen{%
	\footnotesize \textcopyright Copyright 2017 IEEE. Published in the IEEE 2017 International Conference on Image Processing (ICIP 2017), scheduled for 17-20 September 2017 in Beijing, China. Personal use of this material is permitted. However, permission to reprint/republish this material for advertising or promotional purposes or for creating new collective works for resale or redistribution to servers or lists, or to reuse any copyrighted component of this work in other works, must be obtained from the IEEE. Contact: Manager, Copyrights and Permissions / IEEE Service Center / 445 Hoes Lane / P.O. Box 1331 / Piscataway, NJ 08855-1331, USA. Telephone: + Intl. 908-562-3966.}

\newcommand \copyrightnoticeICIP{%
	\begin{tikzpicture}[remember picture,overlay]
	\node[anchor=south,yshift=10pt] at (current page.south) {\fbox{\parbox{\dimexpr\textwidth-\fboxsep-\fboxrule\relax}{ \copyrighttextICIPTwentySeventeen
}}};
	\end{tikzpicture}%
}

\begin{document}
%\ninept
%

\maketitle
\copyrightnoticeICIP	% uncomment for arxiv version

\begin{abstract}
We present a method to perform online Multiple Object Tracking (MOT) of known object categories in monocular video data. Current Tracking-by-Detection MOT approaches build on top of 2D bounding box detections. In contrast, we exploit state-of-the-art instance aware semantic segmentation techniques to compute 2D shape representations of target objects in each frame. We predict position and shape of segmented instances in subsequent frames by exploiting optical flow cues. We define an affinity matrix between instances of subsequent frames which reflects locality and visual similarity. The instance association is solved by applying the Hungarian method. We evaluate different configurations of our algorithm using the MOT 2D 2015 train dataset. The evaluation shows that our tracking approach is able to track objects with high relative motions. In addition, we provide results of our approach on the MOT 2D 2015 test set for comparison with previous works. We achieve a MOTA score of $32.1$. 
% The abstract should contain about 100 to 150 words, and should be identical to the abstract text submitted electronically along with the paper cover sheet. 
\end{abstract}

% Papers with multiple authors and affiliations may require two or more lines for this information. 
% Please note that papers should not be submitted blind; include the authors' names on the PDF.

% To achieve the best rendering both in printed proceedings and electronic proceedings, we strongly encourage you to use Times-Roman font.  

%The first paragraph in each section should not be indented, but all the following paragraphs within the section should be indented.

%In LaTeX, to start a new column (but not a new page) and help balance %the last-page column lengths, you can use \vfill\pagebreak.

% Illustrations must appear within the designated margins. They may span the two columns. 
% If possible, position illustrations at the top of columns, rather
%than in the middle or at the bottom. 
%Caption and number every illustration.
%All halftone illustrations must be clear black and white prints.  %Colors may be used, but they should be selected so as to be readable when printed on a black-only printer.

%Use footnotes sparingly (or not at all!) and place them at the bottom of the column on the page on which they are referenced.

% Below is an example of how to insert images. Delete the ``\vspace'' line,
% uncomment the preceding line ``\centerline...'' and replace ``imageX.ps''
% with a suitable PostScript file name.
% -------------------------------------------------------------------------

\section{Introduction}

\subsection{Motivation}

Tracking-by-Detection is one of the most popular approaches to tackle the problem of online Multiple Object Tracking. The pipeline consists of three stages. In the first stage, objects are independently detected in each frame. Typically, the detections are represented by a two-dimensional bounding box. In the second stage, the detections found in previous and subsequent frames are associated. One way to compute the associations of previous and subsequent objects is by determining a pairwise affinity value. The affinity value may reflect positional information and/or visual similarity. The association is typically solved by applying the Hungarian method to the affinity matrix. In the last stage, previously created tracklets are associated in order to fill the gap between missing detections and to handle occlusions. \\
One possibility to incorporate the position of objects in the corresponding affinity values is by predicting the object position from the current frame to subsequent frames. Typically, this is achieved by using a motion model, e.g. a Kalman Filter, or by exploiting visual cues. Motion based Tracking-by-Detection methods may struggle in scenarios, where camera and object move simultaneously. In this case, the perceived object motion is a superposition of object and camera motion. It is not always possible to describe a superposition of motions adequately using a single motion model. For example, consider the case of a car driving over a speed bump. Suddenly, the position of a person observed from the car experiences a vertical shift. In contrast, optical flow based Tracking-by-Detection does not require the definition of a motion model. Since current optical flow based methods use bounding box representations of the target objects they must deal with non-target-object surfaces contained in the bounding box. Otherwise, occlusions and background structures may influence the quality of the optical flow information.\\
With recently published ConvNets \cite{Dai2015,Li2016} it is possible to segment the two-dimensional shape of instances of known object categories. In contrast to a simple bounding box, this representation has the advantage that it does not contain background structures or parts of other objects. We determine reliable associations of objects in subsequent frames by combining instance aware semantic segmentations and semi-dense optical flow cues. Our approach works online and uses only visual information for object association. Our flow based approach even handles objects with high relative motions, i.e. objects where the observed motion is a superposition of camera and object motion. 

\subsection{Contribution}
% Alternative beginning
% We present a Multiple Object Tracking approach, which exploits instance aware semantic segmentations to compute object associations.
To the best of our knowledge, we present the first instance aware semantic segmentation based Multiple Object Tracking approach. The segmentations allow us to track the two-dimensional shape of objects in subsequent frames on pixel level. We provide a detailed description of the basic elements of our tracking pipeline. We analyze the effectiveness of our method by providing results with different parameter configurations, e.g. different optical flow algorithms, on the MOT 2D 2015 train dataset. In addition, we compare our method with SORT on the MOT 2D 2015 test dataset using detections extracted from instance segmentations. SORT is a open source online MOT tracker, which has shown competitive results using Faster RNN detections.

\subsection{Related Work}

Many state-of-the-art MOT approaches like \cite{Choi15, Xiang2015, Lee16, Yu16} follow the Detection-by-Tracking methodology. \cite{Choi15, Xiang2015} incorporate optical flow algorithms \cite{Lucas1981, Farneback2003} to tackle the problem of object association. Yu et al. \cite{Yu16} achieve state-of-the-art results by training a person specific detector, i.e. they fine-tune the VGG ConvNet \cite{Chatfield14} using several additional training datasets. For online tracking they follow the standard Detection-by-Tracking stages. They use a Kalman filter \cite{Kalman1960} for motion prediction and the Hungarian method \cite{Kuhn1955} to compute associations. Lee et al. \cite{Lee16} present a multi-object tracker based on a Bayesian Filtering framework. Objects are detected using an ensemble of motion detection and object detection. Xiang et al. \cite{Xiang2015} use a Markov Decision Process to perform MOT. Bounding box predictions are computed by measuring the stability of the corresponding optical flow. The quality of the optical flow result is affected by background structures and surfaces of other objects in the bounding box. 
% choi is a single author, therefore no "choi et al"
Choi \cite{Choi15} determines a set of salient points in the input image and computes the corresponding trajectories using the algorithm of Farneback \cite{Farneback2003}. The relative motion of these trajectories w.r.t. a pair of bounding boxes is used to determine an affinity score. Schikora et al. \cite{Schikora11} use optical flow cues to detect moving objects and to track these objects using a particle filter. Milan et al. \cite{Milan2015} use a multi-label conditional random field to assign super pixels to object instances represented by bounding boxes. The superpixels are determined by using color information and optical flow cues. The object associations are computed offline. \\
%To the best of our knowledge, no previous work exploits semantic segmentation instances in the context of Multiple Object Tracking. Semantic segmentation or scene parsing is the task of assigning semantic labels on a pixel level. Current state-of-the-art semantic segmentation pipelines adapt the Fully Convolutional Network architecture proposed by \cite{Long2015}. Recently, \cite{Dai2015, Li2016} extended this ConvNet type to predict instance information in addition to semantic labels. 

\section{Instance Flow Based Tracking}
\label{section:methods}

\newlength{\exampleResultsWidth}
\setlength{\exampleResultsWidth}{1.6in}
\begin{figure*}[!tb]
	\centering
	\subfloat[Instance Segmentations in Image $I_t$]{
		\includegraphics[width=\exampleResultsWidth]{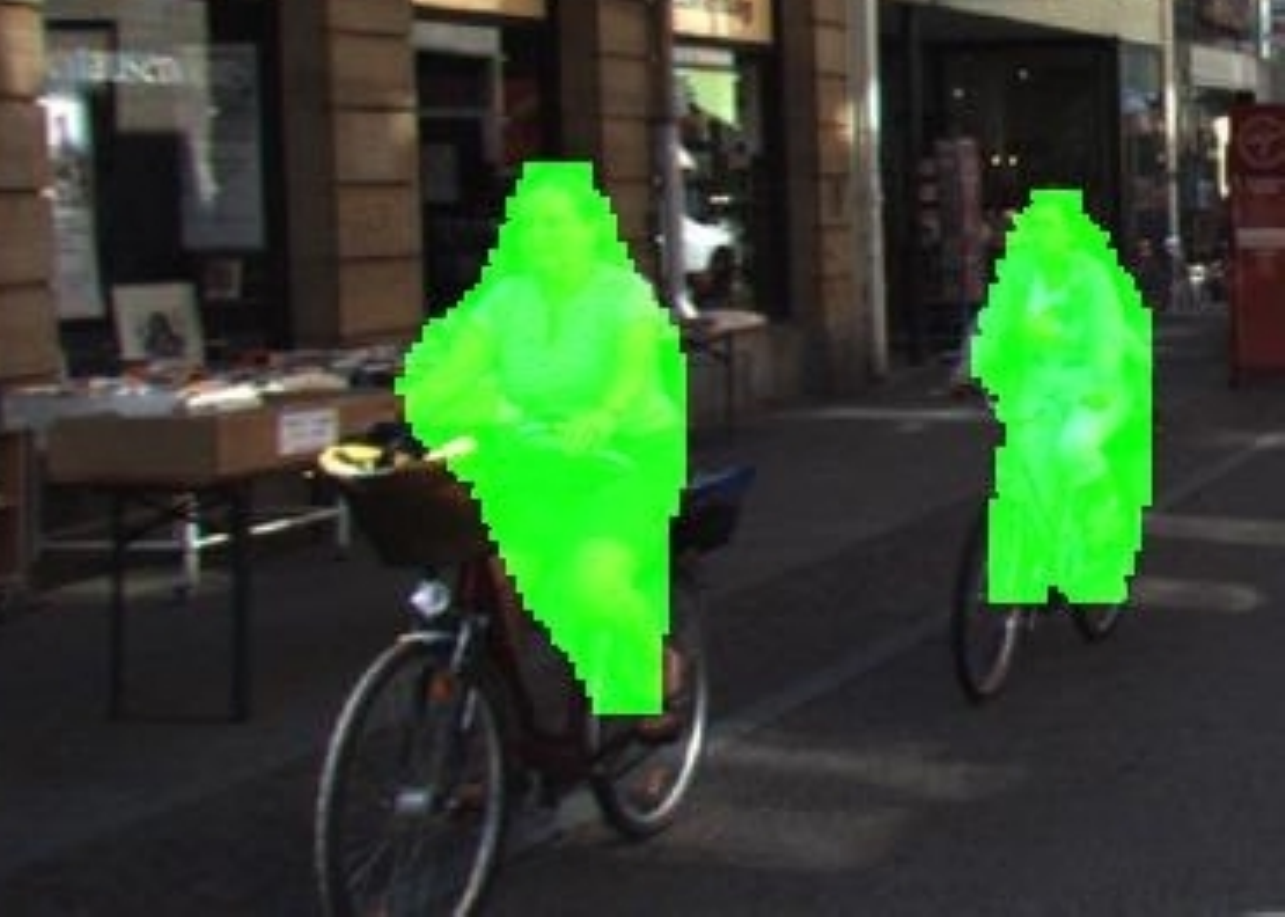}%
		\label{methods:instance_segmentation_i}}
	\hfil
	\subfloat[Predicted Segmentation Instances in Image $I_t+1$]
	{\includegraphics[width=\exampleResultsWidth] {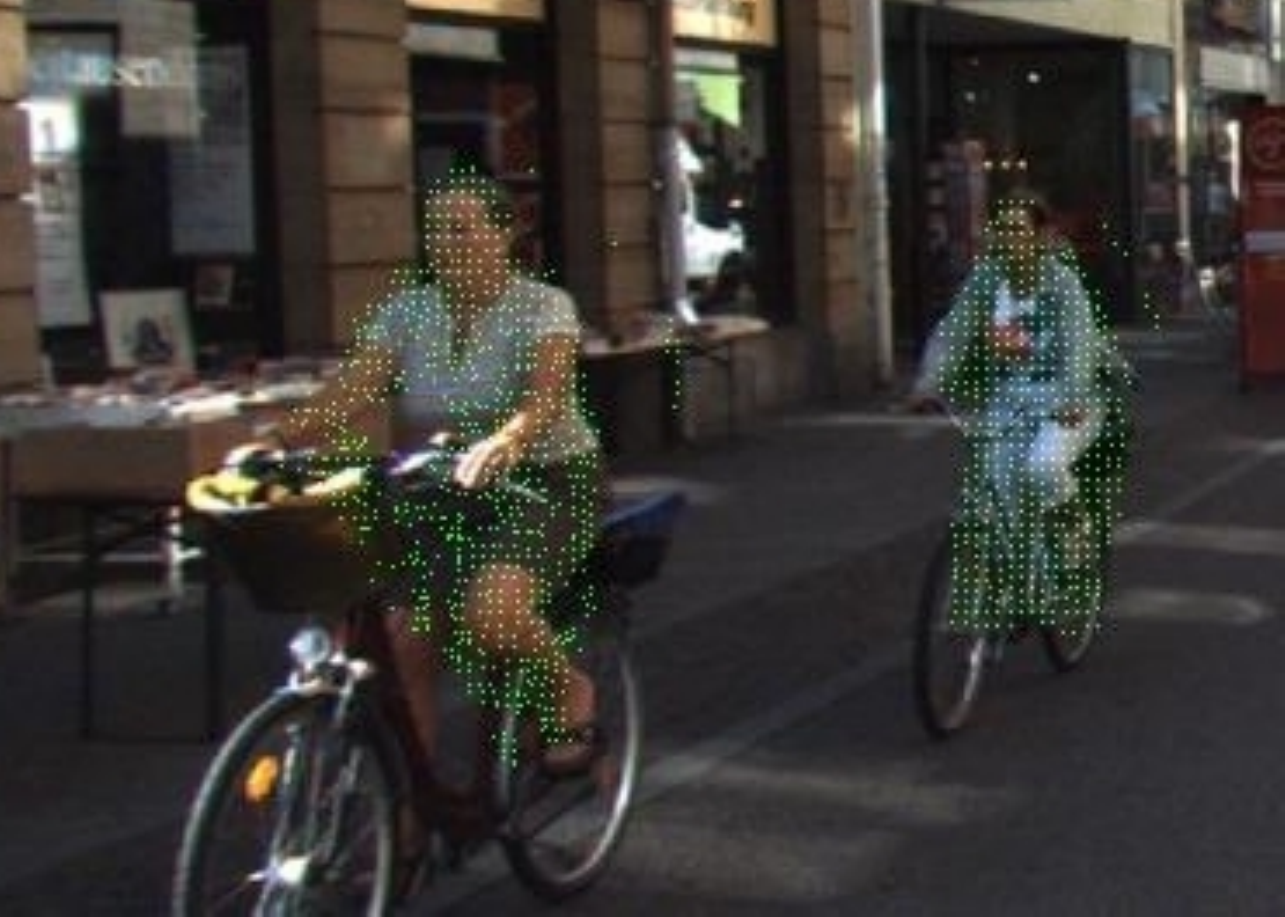}%
		\label{methods:predicted_segmentation_instances}}
	\hfil
	\subfloat[Closed Interpolation of the Predicted Instances in Image $I_t+1$]
	{\includegraphics[width=\exampleResultsWidth] {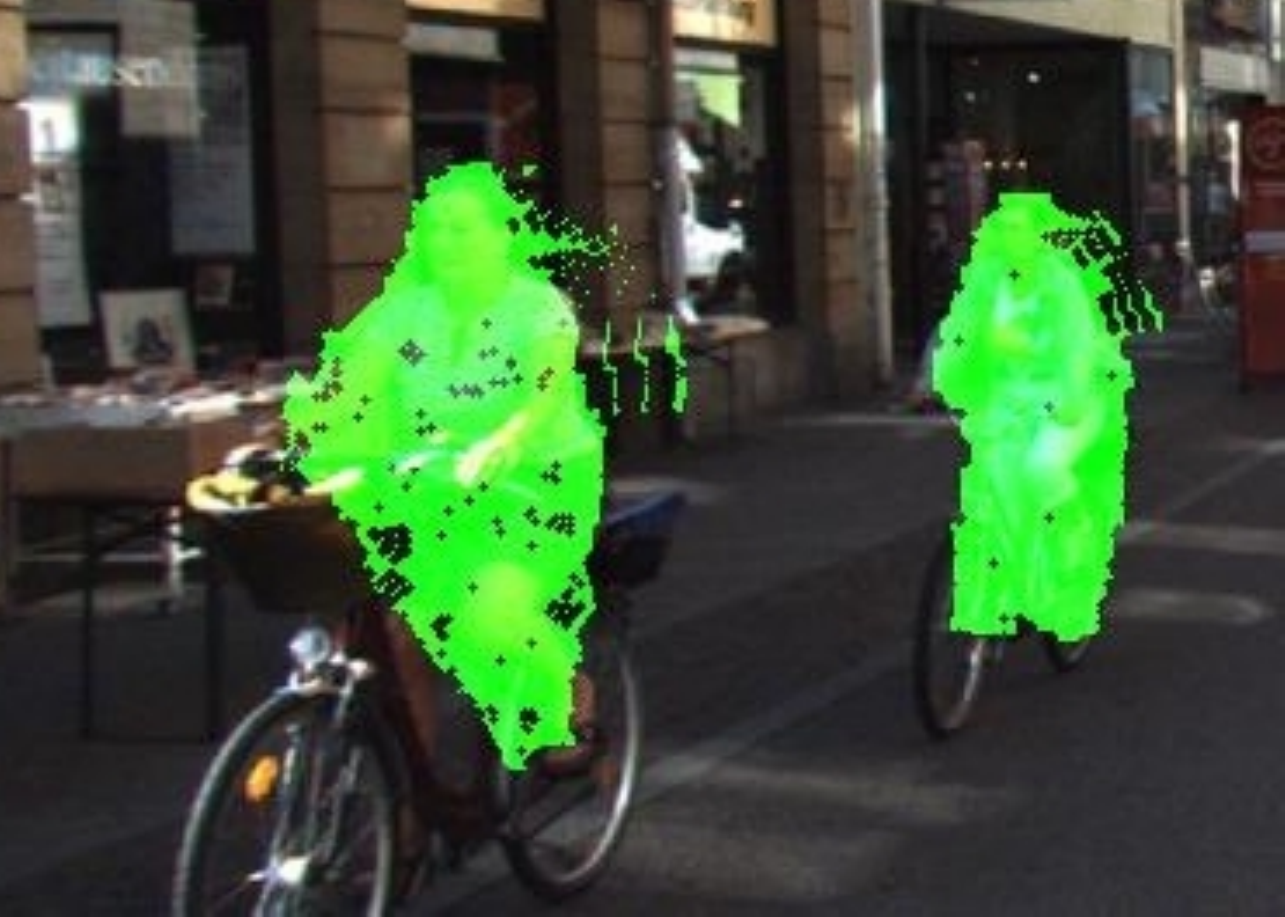}%
			\label{methods:predicted_segmentation_instances_closed}}
	\hfil
	\subfloat[Detected Instances in Image $I_t+1$]
	{\includegraphics[width=\exampleResultsWidth] {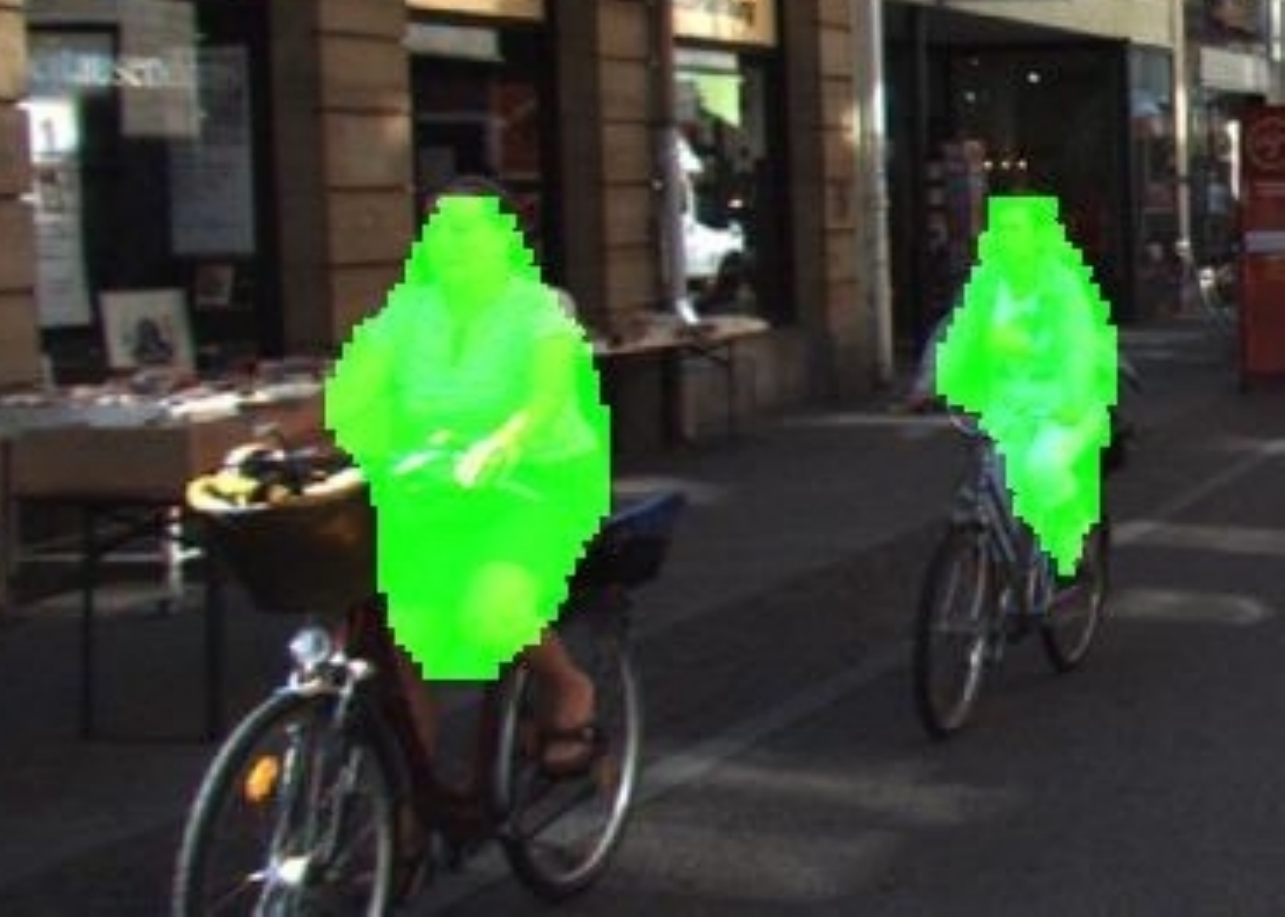}%
		\label{methods:instance_segmentation_i_1}}
	\caption{Instance Flow Prediction using CPM \cite{Hu2016}. Instance flow prediction is best visualized using objects with high velocity, like bicyclists. The offset of corresponding detections is clearly visible. The figure also highlights the importance of accurate segmentations. Otherwise we observe a distortion of the predicted instance segmentation. The figure is best shown in color.}
	\label{methods:basic_concept_instance_flow}
	% Images are taken from the KITTI-19 sequence. Image 35 and 36.
\end{figure*}

\subsection{Terminology}
Let $\vec{I}_t$ denote the t-th image of an ordered sequence with height $h$ and width $w$. Furthermore, let $\vec{I}_{t}(x,y)$ denote the color of pixel position (x,y) in $\vec{I}_t$ with $(x,y) \in \{1,\cdots,w\} \times  \{1,\cdots,h\}$.  \\
Instance aware segmentation systems, like \cite{Dai2015,Li2016}, predict for each pixel position in an input image $\vec{I}_t$ a semantic category label $c$ and a corresponding instance index $i$ according to equation (\ref{eq:instance_segmentation})
\begin{equation}
\label{eq:instance_segmentation}
\begin{split}
\vec{S}_{t}(x,y) = (c, i),
\end{split}
\end{equation}
where $\vec{S}_{t}$ denotes the instance semantic segmentation of image $\vec{I}_t$. \\
Optical Flow or semi-dense matching methods like \cite{Revaud2016, Hu2016} use a pair of subsequent images, denoted as $\vec{I}_t$ and $\vec{I}_{t+o}$, to compute a two-dimensional pixel offset field for pixel positions (x,y). Here, $o$ is the offset of the frame indices in the image sequence. An optical flow algorithm computes the optical flow of an image pair $\vec{F}_{t \rightarrow t+o}(x,y)$ for pixel positions $(x,y)$ such that the similarity in equation (\ref{eq:optical_flow}) is maximized. 
\begin{equation}
\label{eq:optical_flow}
\begin{split}
\vec{I}_{t}(x,y) \simeq \vec{I}_{t+o}(x + \vec{F}_{t \rightarrow t+o}(x,y)[0], \\ y + \vec{F}_{t \rightarrow t+o} (x,y)[1])
\end{split}
\end{equation}
Some optical flow algorithms estimate the optical flow only for a subset of pixels. In this case, there are pixel positions where no optical flow information is available. We will denote the set of pixel positions with valid flow information at time $t$ with $F^{(v)}_{t}$.\\

\subsection{Prediction of Segmentation Instances}
\label{subsection:predictionInstances}

We use the ConvNet presented in \cite{Dai2015} to compute the instance segmentation $\vec{S}_{t}$ for image $\vec{I}_t$. For an instance with index $i$ of the target category $c$ we use $\vec{S}_{t}$ to extract the corresponding set of occupied pixel positions $S_{t,i}$. More formally, we compute $S_{t,i}$ according to equation (\ref{eq:segmentation_instances_target_class})
\begin{equation}
	\label{eq:segmentation_instances_target_class}
	\begin{split}
		S_{t,i} = \{ (x,y)| (x,y) \in \{1,\cdots,w\} \times \{1,\cdots,h\} \\ \land \vec{S}_{t}(x,y) = (c, i) \}.
	\end{split}
\end{equation}
Fig. \ref{methods:instance_segmentation_i} and \ref{methods:instance_segmentation_i_1} show some instance segmentation examples. For subsequent image pairs $\vec{I}_t$ and $\vec{I}_{t+o}$ we compute the optical flow $\vec{F}_{t \rightarrow t+o}$ applying one of the algorithms presented in \cite{Farneback2003, Revaud2016, Hu2016}. This allows us to predict the pixel positions (x,y) contained in $S_{t,i}$ to the next image $\vec{I}_{t+o}$. Fig. \ref{methods:predicted_segmentation_instances} shows the prediction of two instance segmentations using \cite{Hu2016}. We denote the set of predicted pixel positions as $P_{t \rightarrow t+o,i}$ and compute it according to equation (\ref{eq:instance_prediction})
\begin{equation}
\label{eq:instance_prediction}
\begin{split}
P_{t \rightarrow t+o,i} = \{ (x_{p},y_{p}) | (x_{p},y_{p}) = \vec{F}_{t \rightarrow t+o}(x,y) 
\\ \land (x,y) \in F^{(v)}_{t,i}\},
\end{split}
\end{equation}
where $F^{(v)}_{t,i} = S_{t,i} \cap F^{(v)}_{t}$ is the set of valid optical flow positions of instance $i$. If the optical flow algorithm does not provide flow information for each pixel we interpolate the optical flow at positions where no flow information is available. This allows us to compute dense predictions of instance segmentations. We interpolate the optical flow vectors for each instance, separately. To avoid the influence of the optical flow of background structures we use only optical flow vectors of the corresponding instance, i.e. we consider only vectors at pixel positions $F^{(v)}_{t,i}$. We use a linear interpolation of points inside the convex hull of $F^{(v)}_{t,i}$. The optical flow of points lying outside the convex hull is interpolated by using the corresponding nearest neighbor. \\
However, by interpolating optical flow vectors we generate holes and overlaps in the predicted segmentation instance due to optical flow vectors pointing in opposite directions. For example, consider the case of two pixel positions with optical flow vectors pointing away from each other. A gap appears, if points lying in between these two positions are shifted according to an interpolation w.r.t. to their own position. We close these holes by performing a morphological closing operation. An example of a closed interpolation of a predicted segmentation is shown in Fig. \ref{methods:predicted_segmentation_instances_closed}. 

\subsection{Affinity of Objects in Subsequent Frames}
\label{subsection:affinity}
To associate objects visible in image $I_t$ with objects in frame $I_{t+o}$ we compute an affinity score between the corresponding instance segmentations. We define the similarity of an object with index $i$ in frame $I_t$ and object with index $j$ in frame $I_{t+o}$ as the overlap of the intersection of the predicted pixel set $P_{t \rightarrow t+o,i}$ and the pixel set of instance segmentation $S_{t+o,j}$. Note that the number of segmentation instances and the order of the corresponding indices may differ. This formulation of the affinity measure reflects locality and visual similarity. Let $O_{i,j}$ denote the overlap of the prediction $P_{t \rightarrow t+o,i}$ and $S_{t+o,j}$, i.e. $O_{i,j} = \#(P_{t \rightarrow t+o,i} \cap S_{t+o,j})$. Furthermore, let $n_i$ and $n_j$ denote the number of segmentation instances in image $I_t$ and $I_{t+o}$, respectively. We build an affinity matrix $\vec{A}$ using these pairwise overlaps according to equation (\ref{eq:affinity_matrix})
\begin{equation}
\label{eq:affinity_matrix}
\vec{A}_{t \rightarrow t+o} =
\begin{bmatrix}
O_{1,1} 	& \cdots & O_{1,j} 		& \cdots & O_{1,n_j} \\
 \cdots 	& \cdots & \cdots 		& \cdots & \cdots \\
O_{i,1} 	& \cdots & O_{i,j} 		& \cdots & O_{i,n_j} \\
 \cdots 	& \cdots & \cdots 		& \cdots & \cdots \\
O_{n_i,1} 	& \cdots & O_{n_i,j}	& \cdots & O_{n_i,n_j}
\end{bmatrix}
.
\end{equation}

%The matrix has number of instance segmentations recognized in %image $I_t$ rows and number of instance segmentations recognized %in image $I_{t+o}$ columns.

\begin{table*}[!tb]
	\renewcommand{\arraystretch}{1.3}
	\caption{MOT 2D 2015 Benchmark Test Set Evaluation.}
	\label{table:MOT_2D_2015_benchmark_test_set_evaluation}
	\centering
	\begin{tabular}{l|c|ccccccccc}
		Method & $md$ & \textbf{MOTA} $\uparrow$ & MOTP $\uparrow$ & FAF $\downarrow$ & MT $\uparrow$ & ML $\downarrow$ & FP $\downarrow$ & FN $\downarrow$ & ID sw $\downarrow$ & Frag $\downarrow$ \\
		\hline
		FasterRNN+SORT	 	& - & \textbf{33.4} & 72.1 & 1.3 & 11.7\% 	& 30.9\% & 7,318 & 32,615 & 1,001 & 1,764 \\
		MNC+SORT 		  		& - & \textbf{27.5} & 70.5 & 0.5 & 7.5\% 	& 50.9\% & 2,972 & 40,924 & 661   & 1,292\\
		MNC+CPM (ours) 		& 0 & \textbf{30.6} & 71.3 & 0.8 & 10.5\% 	& 34.0\% & 4,863 & 35,325 & 2,459 & 2,953 \\
		MNC+CPM (ours)	& 1 & \textbf{32.1} & 70.9 & 1.1 
		& 13.2\% & 30.1\% & 6,551 & 33,473 & 1,687 & 2,471 \\
	\end{tabular}
\end{table*}

\begin{table*}[!tb]
	\renewcommand{\arraystretch}{1.3}
	\caption{MOT 2015 Benchmark KITTI-13 Evaluation.}
	\label{table:MOT_2015_benchmark_kitti_13_evaluation}
	\centering
	\setlength{\tabcolsep}{4pt} % otherwise table wont fit
	\begin{tabular}{l|c|ccc|cccc|cccc|ccc}
		Method & $md$	& Rcll & Prcn & FAR 
						& GT & MT & PT & ML 
						& FP & FN & IDs & FM 
						& \textbf{MOTA} & MOTP & MOTAL\\
		\hline
		MNC+SORT 				& - & 18.8 & 77.3 & 0.12  
								& 42 & 0 & 14 & 28  
								& 42 & 619 & 3 & 6  
								& \textbf{12.9} & 65.2 & 13.2 \\
		MNC+CPM (ours) 		& 0 & 38.7 & 69.7 & 0.38 
								& 42 & 0 & 32 & 10 
								& 128 & 467 & 25 & 38 
								& \textbf{18.6} & 67.2 & 21.7 \\
		MNC+CPM (ours) 		& 1 & 43.8 & 65.7 & 0.51 
								& 42 & 4 & 32 & 6 
								& 174 & 428 & 14 & 30
								& \textbf{19.2} & 66.7 & 20.8\\
		MNC+DeepMatch (ours) 	& 0 & 38.7 & 69.7 & 0.38
								& 42 & 0 & 32 & 10 
								& 128 & 467 & 25 & 38
								& \textbf{18.6} & 67.2 & 21.7 \\
		MNC+DeepMatch (ours) 	& 1 & 43.8 & 63.8 & 0.55 
								& 42 & 3 & 31 & 8 
								& 188 & 431 & 14 & 29
								& \textbf{16.9} & 66.8 & 18.6  \\
		MNC+PolyExp (ours) 	& 0 & 38.7 & 69.7 & 0.38
								& 42 & 0 & 32 & 10
								& 128 & 467 & 39 & 40
								& \textbf{16.8} & 67.3 & 21.7 \\
		MNC+PolyExp (ours) 	& 1 & 42.7 & 61.2 & 0.61 
							& 42 & 3 & 31 & 8
							& 206 & 437 & 30 & 33
							& \textbf{11.7} & 66.8 & 15.4\\
	\end{tabular}
\end{table*}

\subsection{Online Multiple Object Tracking}

The state of the presented instance flow tracker $T_t$ at time $t$ consists of a set segmentation instances $S_{t,k}$ with unique identifiers $id_{t,k}$ and a counter for the number of missed detections $m_{t,k}$, i.e. $T_t = \{(S_{t,k}, id_{t,k}, m_{t,k}) | k \in \{1, \cdots , n_t\} \}$, where $n_t$ is the number of tracks at time $t$. We initialize this state with the segmentation instances in the first frame (if any). For subsequent frames the tracker state segmentations $S_{t,k}$ are predicted using equation (\ref{eq:instance_prediction}). Let $\vec{I}_t$ denote the previous image and $\vec{I}_{t+1}$ the current image. In order to solve the association of segmentation instances in the tracker state $S_{t,k}$ and segmentations instances $S_{t+1,j}$ found in current image we use the steps described in section \ref{subsection:predictionInstances} and \ref{subsection:affinity} to compute the affinity matrix $\vec{A}_{t \rightarrow t+1}$. We apply the Hungarian Method \cite{Kuhn1955} on $\vec{A}_{t \rightarrow t+1}$, which results in a set of matching index pairs $P_t$. We ensure the validity of an index pair $(k,j) \in P$ by verifying that $\vec{A}_{t \rightarrow t+o}(k,j) > 0$. \\
For all valid index pairs $(k,j) \in P_t$ we update the segmentation instances maintained by the tracker, i.e. we set $S_{t,k} = S_{t+1,j}$, but keep the unique tracklet identifier $id_{t,k}$. We add all non-matching segmentation instances found in image $\vec{I}_{t+1}$ with a new unique identifier to the set of segmentation instances maintained by the tracker. In addition, we remove all non-matching segmentation instances contained in the tracker state, if $m_{t,k} > md$, where $md$ is the number of allowed missing detections. Otherwise, we replace the instance segmentation with a dense prediction of the corresponding pixel positions as described in section \ref{subsection:predictionInstances}. 

% We extend this approach by considering the optical % flow of non-subsequent images. For example, we % compute the optical flow of image $I_t$ and $I_{t+o}$

%\begin{figure}[!tb]               
%	\caption{Instance Flow Multiple Object Tracker}          
%	\label{algorithm:instance_flow_tracker}
%	\begin{algorithmic}
%		\State Compute $S_{0,i}$ with $i = 1, \cdots ,n_i$ 
%		\For{$t = 1, \cdots ,n_t$}
%			\State Compute $P_{t-1 \rightarrow t,i}$ with {$i = 1, \cdots,n_i$}
%			\State Compute $S_{t,j}$ with $j = 1, \cdots,n_j$
%			\State Compute $\vec{M}_{t-1 \rightarrow t}$
%			\State $ip$ = Hungarian($\vec{M}_{t-1 \rightarrow t}$)
%			\State $valid\_ip = \emptyset$
%			\For {$(i,j) \in ip$}
%				\If {$\vec{M}_{t-1 \rightarrow t}[i][j] >0$}
%				\State $valid\_ip = valid\_ip \cup (i,j)$
%				\EndIf
%			\EndFor
%		\EndFor
%	\end{algorithmic}
%\end{figure}

\newlength{\exampleResultsWidthSingleColumn}
\setlength{\exampleResultsWidthSingleColumn}{1.6in}

\section{Evaluation}
\label{section:evaluation}

\begin{figure}[!tb]
	\centering
	\subfloat[Prediction using CPM.]{
		\includegraphics[width=\exampleResultsWidthSingleColumn]{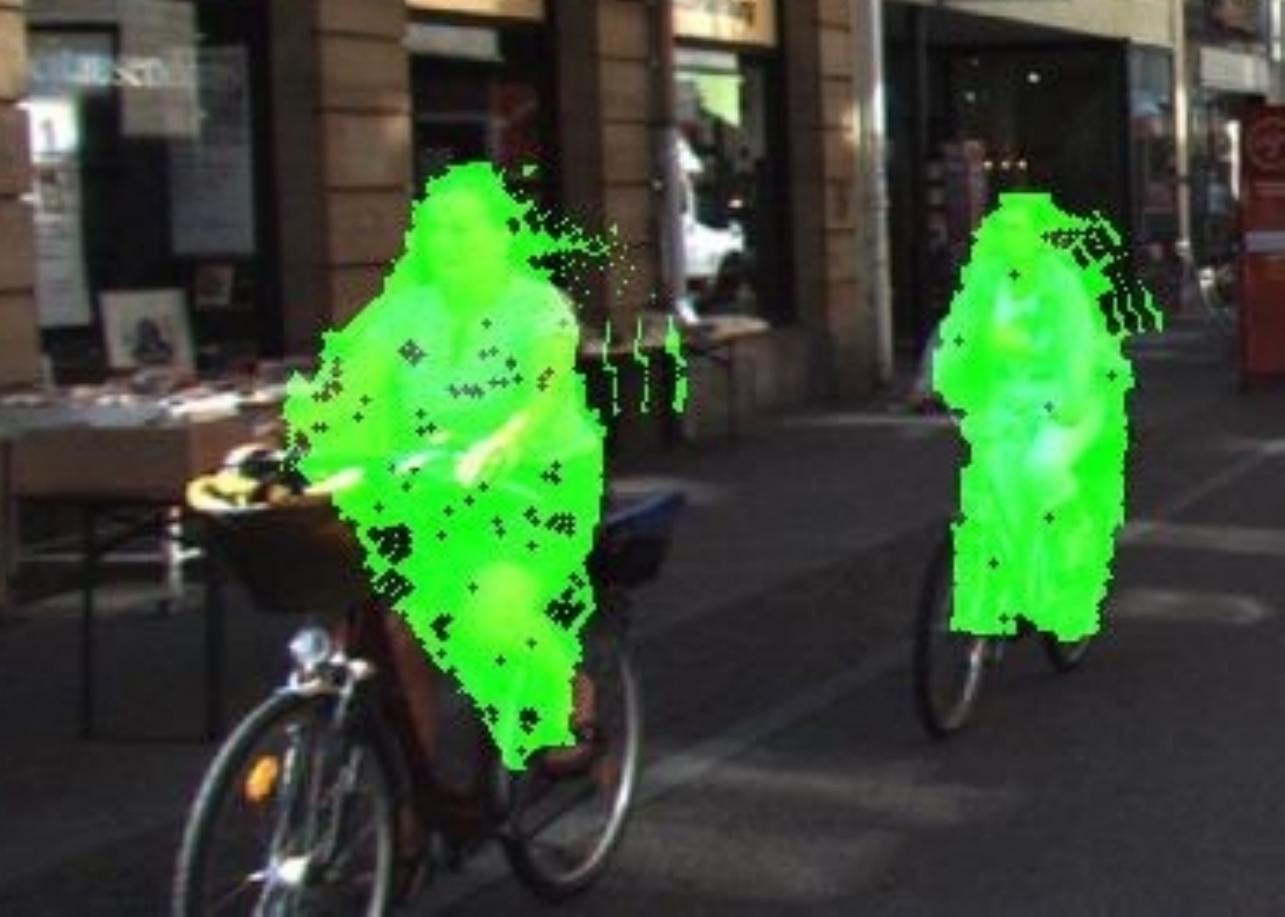}%
		\label{methods:prediction_success}}
	\hfil
	\subfloat[Prediction using PolyExp.]
	{\includegraphics[width=\exampleResultsWidthSingleColumn] {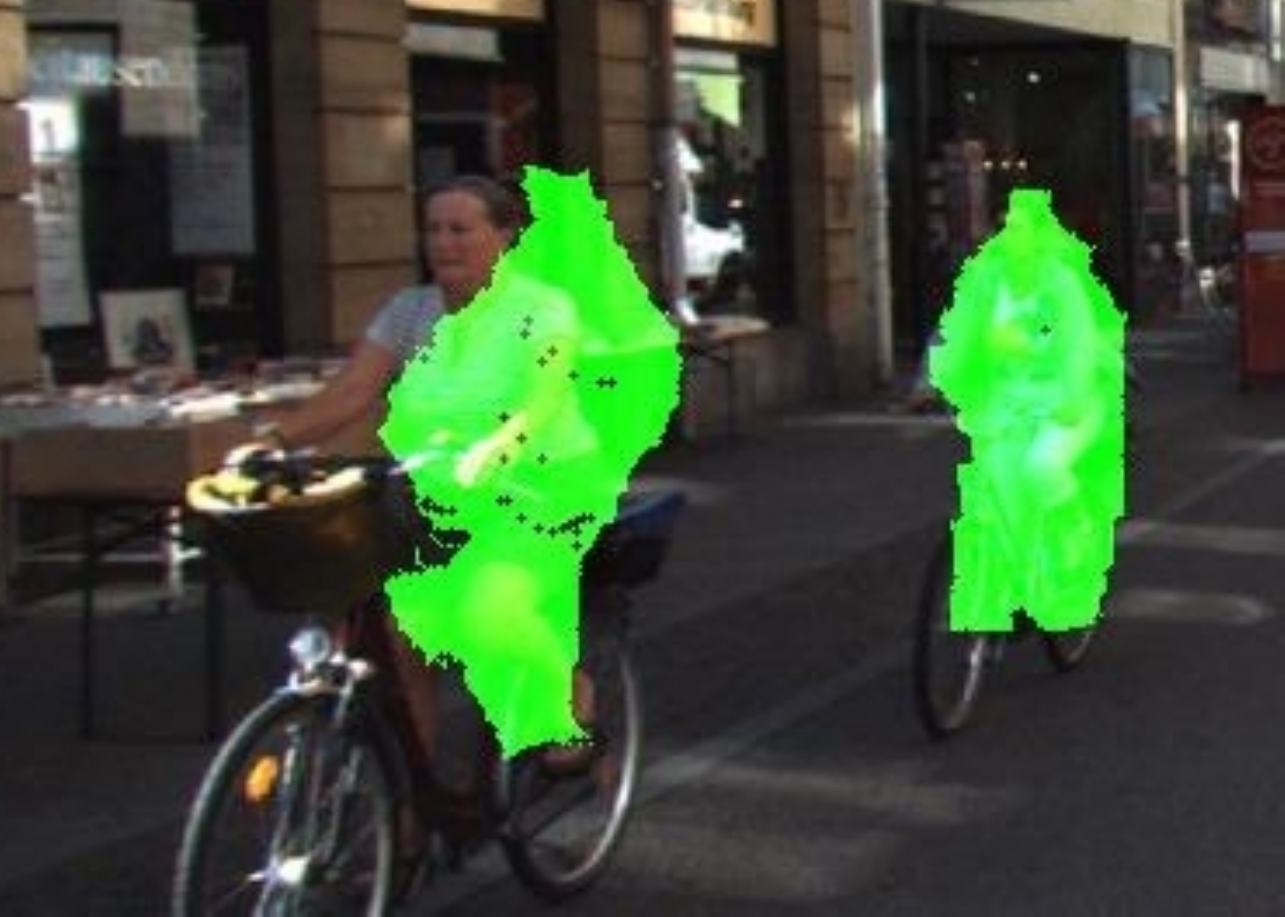}%
		\label{methods:methods:prediction_fail}}
	\caption{Importance of the quality of the optical flow algorithm. The prediction using PolyExp is not correctly shifted.}
	\label{methods:importance_optical_flow_quality}
\end{figure}

We evaluate our Instance Flow based online Multiple Object Tracking approach on the popular MOT dataset \cite{Leal-TaixeS15} using instance aware semantic segmentations computed by \cite{Dai2015} and the optical flow / matching algorithms presented in \cite{Farneback2003}, \cite{Revaud2016} and \cite{Hu2016}. We also analyze the effect of varying the value of $md$. We compare our approach with SORT \cite{Bewley16}, an open source online MOT tracker, which showed competitive results using Faster RNN \cite{Ren15} detections. SORT follows the Tracking-by-Detection pipeline, i.e. Bounding Box detections, a Kalman filter for motion prediction and the Hungarian method for object association. The performance of Tracking-By-Detection approaches is strongly dependent on the quality of detections. By applying SORT on detections derived from instance segmentations we compare the tracking performance without the influence of different detector performances. \\
We use the following combinations in our evaluation:  \textit{FasterRNN +SORT} combines Faster RNN bounding box detections \cite{Ren15} and SORT \cite{Bewley16} tracking. \textit{MNC+SORT} integrates detections extracted from MNC instance segmentations \cite{Dai2015} instead. \textit{MNC+CPM}, \textit{MNC+DeepMatch} and \textit{MNC+PolyExp} use the MNC instance segmentations of \cite{Dai2015} as well as the optical flow of \cite{Hu2016}, the deep matching algorithm of \cite{Revaud2016} and the optical flow of \cite{Farneback2003}, respectively. \\
\textit{MNC+CPM}, \textit{MNC+DeepMatch} and \textit{MNC+PolyExp} achieve similar results on the MOT 2015 training set. A reason for this is the slow motion of camera and pedestrians in most MOT 2015 sequences. In these cases, the quality of object associations is mainly dependent on the segmentation quality. The results of \textit{MNC+CPM} for the test set is shown in table \ref{table:MOT_2D_2015_benchmark_test_set_evaluation}. The biggest difference of the evaluated algorithms in the train dataset is observed in the KITTI-13 sequence, which is the only video captured from a driving platform. In this case, the positions of the objects in image coordinates are strongly affected by the motion of the vehicle, i.e object positions show remarkable shifts between subsequent images. The corresponding results are shown in table \ref{table:MOT_2015_benchmark_kitti_13_evaluation}. In terms of MOTA \textit{MNC+CPM} (with $md = 1$) outperforms \textit{MNC+DeepMatch} as well as \textit{MNC+PolyExp}. This shows the importance of the quality, e.g. density and reliability, of the selected optical flow / matching algorithm. \textit{MNC+DeepMatch} is very sparse and \textit{MNC+PolyExp} and can not handle big object shifts as shown in Fig. \ref{methods:importance_optical_flow_quality}. All optical flow approaches show a higher MOTA score than \textit{MNC+SORT}. This demonstrates the strength of optical flow based approaches in videos with high relative motions of objects. It also shows the difficulty to describe a superposition of motions with a single motion model. We observe, that the number of id switches (IDs) of \textit{MNC+SORT} is significantly lower than the ones of the evaluated optical flow based approaches. This confirms our impression that the used semantic instance segmentation \cite{Dai2015} is unstable. However, we are able to decrease the number of id switches by using dense predictions as instance segmentations in the subsequent frame (e.g. $md = 1$). %Finally, we should note that bounding box detections have a fixed height-width ratio - the rectangles extracted from instance segmentations have not. This also affects the evaluation.

%\begin{figure}[htb]
%	
%	\begin{minipage}[b]{1.0\linewidth}
%		\centering
%		\centerline{\includegraphics[width=8.5cm]{image1}}
%		%  \vspace{2.0cm}
%		\centerline{(a) Result 1}\medskip
%	\end{minipage}
%	%
%	\begin{minipage}[b]{.48\linewidth}
%		\centering
%		\centerline{\includegraphics[width=4.0cm]{image3}}
%		%  \vspace{1.5cm}
%		\centerline{(b) Results 3}\medskip
%	\end{minipage}
%	\hfill
%	\begin{minipage}[b]{0.48\linewidth}
%		\centering
%		\centerline{\includegraphics[width=4.0cm]{image4}}
%		%  \vspace{1.5cm}
%		\centerline{(c) Result 4}\medskip
%	\end{minipage}
%	%
%	\caption{Example of placing a figure with experimental results.}
%	\label{fig:res}
%	%
%\end{figure}
\section{Conclusion}

We presented an online Multiple Object Tracking approach exploiting semantic instance segmentations and optical flow cues. The algorithm is able to track the two-dimensional shape of objects in subsequent frames. We evaluated our approach in the domain of pedestrians. The algorithm shows its benefits while tracking objects with high relative motions. Currently, the tracker only supports basic tracking functionality. In future work, we want to combine our approach with tracker management algorithms to increase its performance by handling occlusions. We demonstrated that semantic instance segmentations are an interesting alternative to conventional bounding box detections.

%Propagation of class probabilities through subsequent images instead of prediction of class instances

% To start a new column (but not a new page) and help balance the last-page
% column length use \vfill\pagebreak.
% -------------------------------------------------------------------------
%\vfill\pagebreak

\bibliographystyle{IEEEbib}
\bibliography{refs}

\end{document}